\documentclass[conference]{IEEEtran}

\usepackage[T1]{fontenc}
\usepackage[utf8]{inputenc}
\usepackage{graphicx}
\usepackage{booktabs}
\usepackage{multirow}
\usepackage{array}
\usepackage{amsmath,amssymb}
\usepackage{url}
\usepackage[hidelinks]{hyperref}
\usepackage{cite}
\usepackage{enumitem}
\usepackage{microtype}

\graphicspath{{images/}}

\title{KITE: A Tri-Modal Transformer Integrating Text, Images, and Knowledge Graphs for Fake News Detection}

\author{\IEEEauthorblockN{Kevin Patel\IEEEauthorrefmark{1}, Shashi Bhushan Jha\IEEEauthorrefmark{1}\textsuperscript{,}\IEEEauthorrefmark{2}}
\IEEEauthorblockA{\IEEEauthorrefmark{1}Department of Computer Science, University of West Florida, Pensacola, FL, USA\\
\IEEEauthorrefmark{2}Corresponding author: Shashi Bhushan Jha (e-mail: sjha@uwf.edu)}}

\begin{document}
\maketitle

\begin{abstract}
Traditional fake news detection methods are falling behind as multimodal misinformation grows more advanced, seamlessly blending deceptive text, manipulated visuals, and factually incorrect claims. Most prior work focuses on text-image fusion or applies external knowledge only as a post-processing step, limiting their ability to detect deeper semantic inconsistencies. In this paper, we introduce KITE (Knowledge-Integrated Text--Image Encoder), a tri-modal fake news detection framework that jointly models textual, visual, and factual knowledge representations. KITE leverages RoBERTa and CLIP for linguistic and visual encoding, while a Graph Attention Network (GAT) processes structured facts retrieved from Wikidata. KITE uses cross-modal attention within a multimodal transformer to integrate text, visual, and knowledge features, helping it understand how each modality relates to one another. Modality-specific confidence scores are generated alongside the final prediction, offering interpretability by indicating which input type most influenced the decision. Evaluations on benchmark datasets demonstrate that KITE outperforms several unimodal and bimodal baselines, particularly in scenarios involving image--text mismatches or contradictions with external knowledge.
\end{abstract}

\begin{IEEEkeywords}
fake news detection, tri-modal fusion, Wikidata fact retrieval, cross-modal transformer, knowledge graph
\end{IEEEkeywords}

\noindent\textbf{Code and dataset availability:} \url{https://github.com/KaiDMML/FakeNewsNet/}

\section{Introduction}
The swift evolution of digital media has fundamentally transformed the way information is created, shared, and consumed. Social platforms and online outlets can share content to millions within seconds, but this speed comes at the expense of verification. As a result, misleading stories commonly referred to as ``fake news'' can spread unchecked, influencing public opinion and even impacting global events. Although early fake news detection efforts primarily relied on textual analysis, visual content, or user interaction features, it has become evident that single-modality approaches are limiting, particularly when malicious actors deliberately alter text or images to bypass detection. Recent advances in multimodal deep learning have shown that combining text and image features can significantly improve classification accuracy~\cite{ref1,ref2,ref4,ref5,ref6}. Models that employ cross-attention mechanisms~\cite{ref2} and unified transformer architectures such as ViLT~\cite{ref7} are highly effective at aligning visual and textual representations, allowing them to detect fake news that would otherwise go unnoticed when analyzing each modality independently.

Knowledge-aware approaches enhance model reliability by incorporating external factual sources such as knowledge graphs and entity-level validation, which help align predictions with verified real-world information. However, most existing frameworks either combine knowledge solely with textual data or incorporate it as a separate verification layer after the initial processing stage, rather than embedding it directly into the core multimodal fusion process. These gaps limit current models' ability to detect sophisticated fake news where misleading language, manipulated images, and factual inconsistencies are intertwined.

To bridge these gaps, this paper proposes KITE (Knowledge-Integrated Text--Image Encoder), a unified architecture that jointly encodes textual content, visual cues, and external factual knowledge. KITE first processes news articles through RoBERTa~\cite{ref8} for text embeddings and CLIP~\cite{ref10} for image embeddings, then retrieves relevant entity facts from Wikidata and encodes them via a Graph Attention Network (GAT). A dynamic cross-modality attention layer aligns and fuses text, image, and knowledge embeddings within a multimodal transformer, allowing the model to detect not only inconsistencies between text and images but also contradictions with external factual knowledge. Finally, a lightweight classification layer generates a binary real/fake prediction along with confidence scores for each modality, providing interpretable insights into the decision-making process.

The main contributions are as follows:
\begin{itemize}[leftmargin=*]
    \item \textbf{Unified multimodal--knowledge fusion.} Text, image, and knowledge-graph tokens attend directly to one another within a single Transformer layer. Unlike prior knowledge-aware models that add external factual knowledge after a separate fusion stage, KITE introduces the knowledge token into the same cross-attention pool.
    \item \textbf{Knowledge-aware GAT encoding.} KITE constructs an article-specific mini knowledge graph from Wikidata and processes it with a two-layer Graph Attention Network whose attention weights are optimized with the fake-news classification objective.
    \item \textbf{Cross-modality attention within a multimodal Transformer.} Inspired by CMA~\cite{ref2}, KITE jointly aligns text, image, and knowledge tokens in the same Transformer layer.
    \item \textbf{Interpretable confidence scoring.} KITE produces modality-specific confidence scores alongside the final prediction to indicate the relative influence of visual, textual, and factual-knowledge streams.
\end{itemize}

\section{Related Work}
\subsection{Single-Modal Methods}
Unimodal approaches to fake news detection analyze a single form of data. The most common domains include textual content, visual representations, and user-interaction features. Chen et al.~\cite{ref12} introduced a rumor-detection model using recurrent neural networks to analyze the evolution of user comments, combined with autoencoders for identifying anomalies in commenting patterns. The model was tested on Sina Weibo data containing 167,731 posts labeled as rumors or non-rumors and 1,501,472 user comments. Although effective within one modality, such methods miss complementary cues from images and other modalities.

Qian et al.~\cite{ref13} developed the Two-level Convolutional Neural Network with User Response Generator (TCNN-URG), which uses article text alongside simulated user responses to identify fake news before it spreads. Qin and Zhang~\cite{ref14} used BERT~\cite{ref15} and a fine-tuning and feature-regularization strategy to improve generalization on BuzzFeedNews and WELFake. Ma et al. applied recurrent models to automatically learn temporal patterns in posts. Segura-Bedmar and Alonso-Bartolome~\cite{ref17} compared unimodal and multimodal approaches on Fakeddit and showed that text-only systems miss useful social and visual cues.

\subsection{Multi-Modal Methods}
As fake-news generation has become more sophisticated, researchers have increasingly integrated text and images. Jin et al.~\cite{ref16} combined recurrent textual/social encoding with VGG-19 visual features. Yu et al.~\cite{ref1} proposed the BERT Multi-Domain and Multi-modal Fusion Network (BMMFN), which combines BERT text embeddings and VGG-19 visual features using co-attention. Jiang et al.~\cite{ref2} proposed Cross-Modal Augmentation (CMA) for few-shot multimodal fake news detection, using CLIP features and cross-attention. Chen et al.~\cite{ref4} introduced a multimodal contrastive-learning approach involving text, images, and propagation networks. Wu et al.~\cite{ref5} proposed Balanced Multi-modal Learning with Hierarchical Fusion (BMLHF), which dynamically balances text and image contributions. Qu et al.~\cite{ref6} proposed QMFND, a quantum multimodal fusion-based framework. Although these approaches improve performance by combining modalities, they generally lack explicit external factual grounding.

\subsection{Knowledge-Aware Methods}
External knowledge can ground predictions in factual evidence. Knowledge-aware methods incorporate knowledge bases and semantic frameworks to reduce dependence on content analysis alone~\cite{ref9,ref18,ref19}. CompareNet~\cite{ref20} constructs document graphs that compare news text against external knowledge. KAN~\cite{ref21} integrates semantic and knowledge-level representations using attention. KMAGCN~\cite{ref11} combines text, image, and knowledge in a graph-based representation using graph convolutional networks. Hu et al.~\cite{ref22} proposed a matching-aware co-attention mechanism for aligning textual claims and visual content. Thilagam and Karur~\cite{ref23} used multilayer perceptrons to assess knowledge-base triples. Nair et al.~\cite{ref24} constructed a knowledge base with subject--predicate--object triples, sentiment polarity, occurrence counts, and topic-model indicators.

\subsection{Research Gap}
Despite notable advancements, existing fake news detection models still face limitations. Unimodal architectures overlook complementary cues from other modalities. Multimodal methods improve upon unimodal systems by combining textual and visual data, but often depend on early concatenation or relatively simple attention mechanisms. Knowledge-aware models frequently use external knowledge as supplementary embeddings or as a post-processing component rather than integrating it into the core fusion process.

To address these limitations, KITE introduces a unified tri-modal fusion strategy. Text embeddings from RoBERTa, visual embeddings from CLIP, and factual knowledge embeddings from a GAT are combined in a lightweight cross-modal Transformer. By aligning text, image, and external knowledge representations simultaneously, KITE is designed to detect inconsistencies between multimodal content and external facts.

\section{Methodology}
\begin{figure*}[t]
    \centering
    \includegraphics[width=0.95\textwidth]{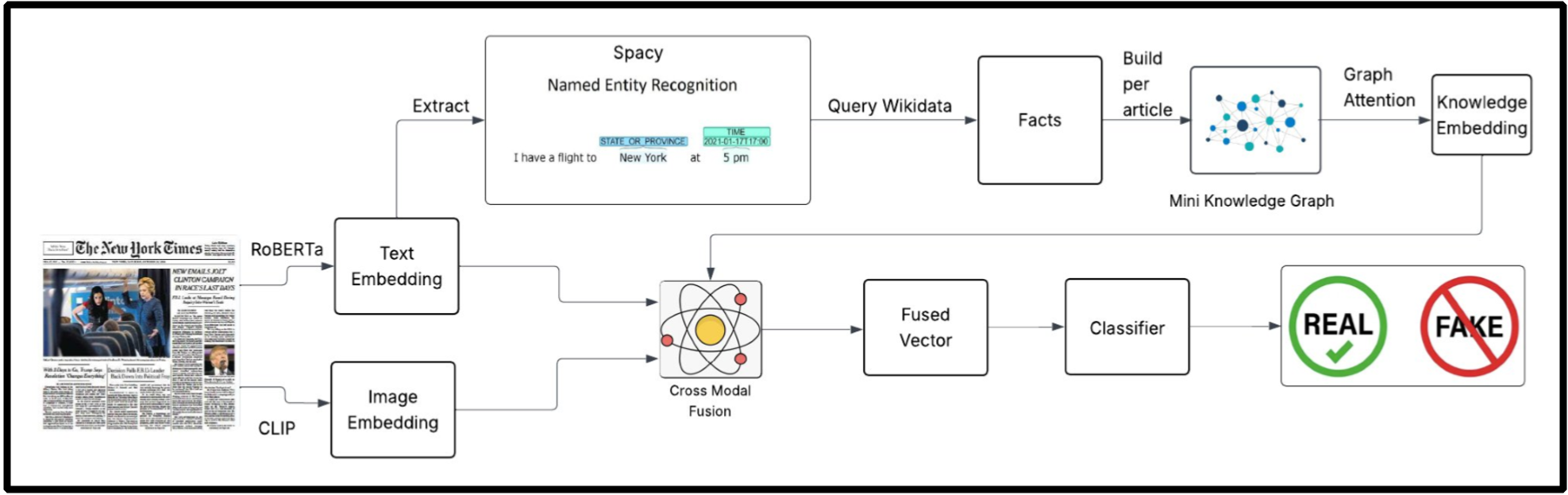}
    \caption{Architecture of the KITE model.}
    \label{fig:kite}
\end{figure*}

This section describes the Knowledge-Integrated Text--Image Encoder (KITE). The architecture has four main stages: input processing for text and images, integration of external factual knowledge, multimodal fusion, and final classification. Figure~\ref{fig:kite} provides an overview.

\subsection{Input Processing (Text and Images)}
The initial stage of KITE encodes textual and visual content into fixed-length embeddings.

\subsubsection{Text Encoding with RoBERTa}
Textual input is processed using RoBERTa~\cite{ref8}, a robustly optimized version of BERT. The input text is tokenized and embeddings from the final layer are used as textual features. The implementation uses the pre-trained \texttt{roberta-base} model through the Hugging Face Transformers library.

\subsubsection{Image Encoding with CLIP}
Visual content is encoded using the Contrastive Language--Image Pretraining (CLIP) model~\cite{ref10}. KITE uses the pretrained \texttt{clip-vit-base-patch32} model through Hugging Face Transformers to obtain image features.

\subsection{Knowledge-Integrated Text--Image Encoder Architecture}
\subsubsection{Enhanced Cross-Modality Fusion}
KITE replaces simple concatenation with a Cross-Modal Transformer (CMT) inspired by prior cross-modal attention work~\cite{ref2,ref25}. Each modality is treated as a learned token and projected to a common hidden size. A Transformer encoder then allows the text, image, and knowledge representations to interact through self-attention. The pooled representation is forwarded to the classifier head.

\subsubsection{Fact-Checking Module}
The fact-checking module is inspired by knowledge-aware multimodal pre-training~\cite{ref9} and contains three steps:
\begin{itemize}[leftmargin=*]
    \item \textbf{Named Entity Recognition (NER):} persons, organizations, and locations are extracted from article text using spaCy.
    \item \textbf{Knowledge Retrieval:} extracted entities are used to query Wikidata and retrieve factual information.
    \item \textbf{Knowledge Encoder (GAT):} the resulting entity--fact graph is encoded with a two-layer Graph Attention Network. The final graph readout forms a 256-dimensional knowledge embedding.
\end{itemize}

\subsubsection{Fusion with Knowledge}
Text, image, and factual knowledge are represented as modality tokens and passed through the lightweight Cross-Modal Transformer. Self-attention allows each modality token to interact with the others. The pooled output is then passed to the classifier and confidence heads.

\subsubsection{Output Classification}
The fused representation is passed through a linear layer with a sigmoid activation~\cite{ref26} to produce the probability that an article is fake. Three modality-specific heads produce confidence scores for the text, image, and knowledge streams. These heads are trained jointly with the main classifier under the same focal-loss objective.

\section{Experiments}
\subsection{Datasets}
KITE is evaluated using GossipCop and PolitiFact from the FakeNewsNet benchmark suite. GossipCop focuses primarily on entertainment and celebrity news, while PolitiFact contains political news and fact-checked claims. Both datasets contain article text and associated images after preprocessing/filtering. Table~\ref{tab:dataset} summarizes the dataset statistics reported in the source manuscript.

\begin{table}[t]
\centering
\caption{Statistics of news datasets.}
\label{tab:dataset}
\resizebox{\columnwidth}{!}{%
\begin{tabular}{lrrrrr}
\toprule
Dataset & Total & Train & Test & Real (\%) & Fake (\%)\\
\midrule
GossipCop & 8448 & 6758 & 1690 & 54.57 & 45.43\\
PolitiFact & 325 & 260 & 65 & 52.00 & 48.00\\
\bottomrule
\end{tabular}}
\end{table}

\subsection{Experimental Setup}
The GossipCop dataset was split into 6,758 training and 1,690 testing samples, while PolitiFact was partitioned into 260 training and 65 testing samples. KITE was trained for 10 epochs with batch size 32 and learning rate $1\times10^{-4}$. AdamW was used with weight decay $1\times10^{-3}$. A linear learning-rate scheduler with warm-up was applied: the learning rate increased linearly during the initial 10\% of training steps and then decayed gradually.

The experiments were conducted on CPU hardware. The implementation used Python, PyTorch 2.1.0, Hugging Face Transformers 4.35.2, PyTorch Geometric 2.5.0, and spaCy 3.8.4.

Performance was evaluated using accuracy, precision, recall, and F1 score:
\begin{align}
\mathrm{Accuracy} &= \frac{TP+TN}{TP+TN+FP+FN},\\
\mathrm{Precision} &= \frac{TP}{TP+FP},\\
\mathrm{Recall} &= \frac{TP}{TP+FN},\\
F_1 &= 2\frac{\mathrm{Precision}\cdot\mathrm{Recall}}{\mathrm{Precision}+\mathrm{Recall}}.
\end{align}
The source manuscript states that reported results are averaged across multiple runs.

\subsection{Baselines}
\subsubsection{Unimodal Baselines}
The source manuscript compares KITE with reported results for BERT, XLNet, text CNN, and image CNN baselines on GossipCop and PolitiFact. Table~\ref{tab:unimodal} reproduces those values.

\begin{table*}[t]
\centering
\caption{Baseline unimodal model performance.}
\label{tab:unimodal}
\begin{tabular}{llrrrr}
\toprule
Dataset & Model & Accuracy & Recall & Precision & F1\\
\midrule
GossipCop & BERT & 0.871 & 0.914 & 0.924 & 0.919\\
GossipCop & XLNet & 0.884 & 0.943 & 0.917 & 0.929\\
GossipCop & CNN (Text) & 0.806 & 0.643 & 0.882 & 0.744\\
GossipCop & CNN (Image) & 0.802 & 0.929 & 0.842 & 0.883\\
PolitiFact & BERT & 0.843 & 0.696 & 0.904 & 0.786\\
PolitiFact & XLNet & 0.847 & 0.704 & 0.905 & 0.792\\
PolitiFact & CNN (Text) & 0.776 & 0.614 & 0.857 & 0.715\\
PolitiFact & CNN (Image) & 0.692 & 0.733 & 0.821 & 0.774\\
\bottomrule
\end{tabular}
\end{table*}

\subsubsection{Multimodal and Knowledge-Aware Baselines}
The comparison includes MCAN~\cite{ref27}, MCAN-A~\cite{ref27}, SpotFake+~\cite{ref28}, QMFND~\cite{ref6}, and KAN~\cite{ref25}. Table~\ref{tab:multimodal} reproduces the performance values reported in the manuscript.

\begin{table*}[t]
\centering
\caption{Baseline multimodal/knowledge-aware model performance.}
\label{tab:multimodal}
\begin{tabular}{llrrrr}
\toprule
Dataset & Model & Accuracy & Recall & Precision & F1\\
\midrule
GossipCop & MCAN & 0.869 & 0.806 & 0.890 & 0.846\\
GossipCop & MCAN-A & 0.851 & 0.859 & 0.877 & 0.824\\
GossipCop & SpotFake+ & 0.839 & 0.799 & 0.853 & 0.825\\
GossipCop & QMFND & 0.879 & 0.958 & 0.899 & 0.928\\
GossipCop & KAN & 0.777 & 0.770 & 0.776 & 0.771\\
PolitiFact & MCAN & 0.846 & 0.829 & 0.851 & 0.840\\
PolitiFact & MCAN-A & 0.809 & 0.758 & 0.827 & 0.792\\
PolitiFact & SpotFake+ & 0.789 & 0.753 & 0.803 & 0.777\\
PolitiFact & QMFND & 0.846 & 0.853 & 0.927 & 0.888\\
PolitiFact & KAN & 0.920 & 0.850 & 0.869 & 0.854\\
\bottomrule
\end{tabular}
\end{table*}

\subsection{Results}
Tables~\ref{tab:gossip} and~\ref{tab:politifact} compare KITE against the baselines reported in the source manuscript.

\begin{table*}[t]
\centering
\caption{GossipCop: comparison of KITE with existing models.}
\label{tab:gossip}
\resizebox{\textwidth}{!}{%
\begin{tabular}{lrrrrrrrrrr}
\toprule
Metric & BERT & XLNet & CNN-T & CNN-I & MCAN & MCAN-A & SpotFake+ & QMFND & KAN & KITE\\
\midrule
Accuracy & 0.871 & 0.884 & 0.806 & 0.802 & 0.869 & 0.851 & 0.839 & 0.879 & 0.777 & \textbf{0.885}\\
Recall & 0.914 & 0.943 & 0.643 & 0.929 & 0.806 & 0.859 & 0.799 & 0.958 & 0.770 & \textbf{0.959}\\
Precision & 0.924 & 0.917 & 0.882 & 0.842 & 0.890 & 0.877 & 0.853 & 0.899 & 0.776 & \textbf{0.926}\\
F1 & 0.919 & 0.929 & 0.744 & 0.883 & 0.846 & 0.824 & 0.825 & 0.928 & 0.771 & \textbf{0.930}\\
\bottomrule
\end{tabular}}
\end{table*}

\begin{table*}[t]
\centering
\caption{PolitiFact: comparison of KITE with existing models.}
\label{tab:politifact}
\resizebox{\textwidth}{!}{%
\begin{tabular}{lrrrrrrrrrr}
\toprule
Metric & BERT & XLNet & CNN-T & CNN-I & MCAN & MCAN-A & SpotFake+ & QMFND & KAN & KITE\\
\midrule
Accuracy & 0.843 & 0.847 & 0.776 & 0.692 & 0.846 & 0.809 & 0.789 & 0.846 & 0.859 & \textbf{0.868}\\
Recall & 0.696 & 0.704 & 0.614 & 0.733 & 0.829 & 0.758 & 0.753 & 0.853 & 0.850 & \textbf{0.879}\\
Precision & 0.904 & 0.905 & 0.857 & 0.821 & 0.851 & 0.827 & 0.803 & 0.927 & 0.869 & \textbf{0.935}\\
F1 & 0.786 & 0.792 & 0.715 & 0.774 & 0.840 & 0.792 & 0.777 & 0.888 & 0.854 & \textbf{0.889}\\
\bottomrule
\end{tabular}}
\end{table*}

The source manuscript attributes KITE's gains to three architectural elements: (1) tri-modal fusion through a cross-modal Transformer, (2) graph-attention-based knowledge encoding, and (3) end-to-end fine-tuning of the final layers of RoBERTa and CLIP. The tri-modal design allows text, image, and knowledge representations to interact jointly; the GAT provides article-specific factual grounding; and selective fine-tuning adapts pretrained encoders to the fake-news domain.

\section{Discussion}
\subsection{Common Failure Causes}
Two recurring failure patterns are identified. First, textual ambiguity such as sarcasm, satire, or vague phrasing can lead to incorrect classifications. Second, low-resolution, generic, irrelevant, or ambiguous images can reduce the quality of CLIP embeddings and introduce noise into multimodal fusion.

\subsection{Dataset Limitations}
The FakeNewsNet data used in the study contain incomplete or missing multimodal content. Many articles include text without corresponding images. The source manuscript reports that manual filtering was used to remove text-only articles, reducing the usable dataset size and potentially constraining generalizability.

\subsection{Computational Challenges}
Training was conducted on CPU hardware. Wikidata retrieval introduced additional latency, and the source manuscript reports training times of approximately 20 hours. This may become a scalability limitation for larger datasets or repeated experimentation. 

\section{Conclusion}
This paper introduces KITE, a tri-modal architecture for fake news detection that fuses textual content, visual signals, and structured external knowledge. By combining cross-modal attention, graph-based knowledge encoding, and selective fine-tuning of RoBERTa and CLIP, KITE is designed to detect both text--image mismatches and contradictions with external facts. The reported results on GossipCop and PolitiFact show competitive performance relative to the selected unimodal and multimodal baselines.

Future work includes automatic data augmentation for low-resource settings, cross-lingual misinformation detection, and deception-aware learning strategies intended to improve robustness against intentionally misleading content.


\begin{thebibliography}{28}

\bibitem{ref1} K. Yu, S. Jiao, and Z. Ma, ``Fake news detection based on BERT multi-domain and multi-modal fusion network,'' \emph{Computer Vision and Image Understanding}, 2025, Art. no. 104301.

\bibitem{ref2} Y. Jiang, T. Wang, and X. Xu, ``Cross-modal augmentation for few-shot multimodal fake news detection,'' \emph{Engineering Applications of Artificial Intelligence}, 2025.

\bibitem{ref3} A. Radford \emph{et al.}, ``Learning transferable visual models from natural language supervision,'' in \emph{Proc. 38th Int. Conf. Machine Learning}, PMLR, 2021, pp. 8748--8763.

\bibitem{ref4} H. Chen, H. Wang, Z. Liu, \emph{et al.}, ``Multi-modal robustness fake news detection with crossmodal and propagation network contrastive learning,'' \emph{Knowledge-Based Systems}, vol. 309, 2025, Art. no. 112800.

\bibitem{ref5} F. Wu, S. Chen, and G. Gao, ``Balanced multi-modal learning with hierarchical fusion for fake news detection,'' \emph{Pattern Recognition}, 2025.

\bibitem{ref6} Z. Qu, Y. Meng, G. Muhammad, and P. Tiwari, ``QMFND: A quantum multimodal fusion-based fake news detection model for social media,'' \emph{Information Fusion}, vol. 104, 2024, Art. no. 102172.

\bibitem{ref7} W. Kim, B. Son, and I. Kim, ``ViLT: Vision-and-language transformer without convolution or region supervision,'' in \emph{Proc. Int. Conf. Machine Learning}, PMLR, 2021, pp. 5583--5594.

\bibitem{ref8} Y. Liu \emph{et al.}, ``RoBERTa: A robustly optimized BERT pretraining approach,'' arXiv:1907.11692, 2019.

\bibitem{ref9} L. Zhang, X. Zhang, Z. Zhou, X. Zhang, P. S. Yu, and C. Li, ``Knowledge-aware multimodal pre-training for fake news detection,'' \emph{Information Fusion}, 2024, Art. no. 102715, doi: 10.1016/j.inffus.2024.102715.

\bibitem{ref10} A. Radford \emph{et al.}, ``Learning transferable visual models from natural language supervision,'' in \emph{Proc. Int. Conf. Machine Learning}, PMLR, 2021, pp. 8748--8763.

\bibitem{ref11} S. Qian, J. Hu, Q. Fang, and C. Xu, ``Knowledge-aware multi-modal adaptive graph convolutional networks for fake news detection,'' \emph{ACM Trans. Multimedia Computing, Communications, and Applications}, vol. 17, no. 3, 2021, Art. no. 98.

\bibitem{ref12} W. Chen, Y. Zhang, C. K. Yeo, C. T. Lau, and B. S. Lee, ``Unsupervised rumor detection based on users' behaviors using neural networks,'' \emph{Pattern Recognition Letters}, vol. 105, pp. 226--233, 2018.

\bibitem{ref13} F. Qian, C. Y. Gong, K. Sharma, and Y. Liu, ``Neural user response generator: Fake news detection with collective user intelligence,'' in \emph{Proc. IJCAI}, 2018, pp. 3834--3840.

\bibitem{ref14} S. Qin and M. Zhang, ``Boosting generalization of fine-tuning BERT for fake news detection,'' 2024. [Online]. Available: \url{https://www.sciencedirect.com/science/article/pii/S0306457324001055}

\bibitem{ref15} J. Devlin, M.-W. Chang, K. Lee, and K. Toutanova, ``BERT: Pre-training of deep bidirectional transformers for language understanding,'' in \emph{Proc. NAACL-HLT}, 2019, pp. 4171--4186.

\bibitem{ref16} Z. Jin, J. Cao, H. Guo, Y. Zhang, and J. Luo, ``Multimodal fusion with recurrent neural networks for rumor detection on microblogs,'' in \emph{Proc. 25th ACM Int. Conf. Multimedia}, 2017, pp. 795--816.

\bibitem{ref17} I. Segura-Bedmar and S. Alonso-Bartolome, ``Multimodal fake news detection,'' \emph{Information}, vol. 13, no. 6, p. 284, 2022, doi: 10.3390/info13060284.

\bibitem{ref18} X. Chen, X. Huang, Q. Gao, L. Huang, and G. Liu, ``Enhancing text-centric fake news detection via external knowledge distillation from LLMs,'' \emph{Neural Networks}, vol. 187, 2025, Art. no. 107377.

\bibitem{ref19} S. K. and P. S. Thilagam, ``Multi-layer perceptron based fake news classification using knowledge base triples,'' \emph{Applied Intelligence}, vol. 53, 2023, pp. 6276--6287, doi: 10.1007/s10489-02203627-9.

\bibitem{ref20} L. Hu, T. Yang, L. Zhang, W. Zhong, D. Tang, C. Shi, N. Duan, and M. Zhou, ``Compare to the knowledge: Graph neural fake news detection with external knowledge,'' in \emph{Proc. ACL}, 2021, doi: 10.18653/v1/2021.acl-long.62.

\bibitem{ref21} D. Dun, L. Hu, L. Zhang, and C. Shi, ``KAN: Knowledge-aware attention network for fake news detection,'' in \emph{Proc. AAAI Conf. Artificial Intelligence}, vol. 35, no. 1, 2021, pp. 81--89.

\bibitem{ref22} L. Hu, Z. Zhao, W. Qi, X. Song, and L. Nie, ``Multimodal matching-aware co-attention networks with mutual knowledge distillation for fake news detection,'' \emph{Information Sciences}, vol. 664, 2024, Art. no. 120310.

\bibitem{ref23} P. S. Thilagam and S. Karur, ``Multi-layer perceptron-based fake news classification using knowledge base triples,'' \emph{Applied Intelligence}, vol. 53, no. 6, pp. 7129--7143, 2023, doi: 10.1007/s10489-022-03627-9.

\bibitem{ref24} V. Nair, J. Pareek, and S. Bhatt, ``A knowledge-based deep learning approach for automatic fake news detection using BERT on Twitter,'' \emph{Procedia Computer Science}, vol. 235, pp. 1870--1882, 2024.

\bibitem{ref25} W. Zhang, Q. Tan, P. Li, Q. Zhang, and R. Wang, ``Cross-modal transformer with language query for referring image segmentation,'' \emph{Neurocomputing}, vol. 536, pp. 191--205, 2023.

\bibitem{ref26} S. Venkatesh, R. Sindhu, and V. Arunachalam, ``Hardware efficient approximate sigmoid activation function for classifying features around zero,'' \emph{Integration}, vol. 103, 2025, Art. no. 102421.

\bibitem{ref27} Y. Wu, P. Zhan, Y. Zhang, L. Wang, and Z. Xu, ``Multimodal fusion with co-attention networks for fake news detection,'' in \emph{Findings of ACL-IJCNLP}, 2021, pp. 2560--2569.

\bibitem{ref28} S. Singhal, A. Kabra, M. Sharma, R. R. Shah, T. Chakraborty, and P. Kumaraguru, ``SpotFake+: A multimodal framework for fake news detection via transfer learning,'' in \emph{Proc. AAAI Conf. Artificial Intelligence}, vol. 34, no. 10, 2020, pp. 13915--13916.

\end{thebibliography}
\end{document}